# The Innovation-to-Occupations Ontology: Linking Business Transformation Initiatives to Occupations and Skills

### Full research paper

**Daniela Elia**
University of Technology Sydney
Sydney, Australia
Email: daniela.elia@student.uts.edu.au

**Fang Chen**
University of Technology Sydney
Sydney, Australia
Email: fang.chen@uts.edu.au

**Didar Zowghi**
Data61, CSIRO
Sydney, Australia
Email: didar.zowghi@data61.csiro.au

**Marian-Andrei Rizoiu**
University of Technology Sydney
Sydney, Australia
Email: marian-andrei.rizoiu@uts.edu.au

## Abstract

The fast adoption of new technologies forces companies to continuously adapt their operations making it harder to predict workforce requirements. Several recent studies have attempted to predict the emergence of new roles and skills in the labour market from online job ads. This paper aims to present a novel ontology linking business transformation initiatives to occupations and an approach to automatically populating it by leveraging embeddings extracted from job ads and Wikipedia pages on business transformation and emerging technologies topics. To our knowledge, no previous research explicitly links business transformation initiatives, like the adoption of new technologies or the entry into new markets, to the roles needed. Our approach successfully matches occupations to transformation initiatives under ten different scenarios, five linked to technology adoption and five related to business. This framework presents an innovative approach to guide enterprises and educational institutions on the workforce requirements for specific business transformation initiatives.

# 1 Introduction

Fast technological advancements have caused rapid changes in the workforce composition of organisations. Over the last few years, the move to hybrid work has accelerated the demand for digital products and services (Santoso et al. 2021; Lokuge and Sedera 2020) and increased the investment in technologies that enable hybrid capabilities -- such as collaborative work management solutions, desktop-as-a-service applications, and virtual meeting solutions. More recently, the interest and commercial adoption of generative AI technologies has changed how companies operate even faster. This rapid adoption of emerging technologies is bound to alter how companies function, creating job dislocation and demand for new roles and generating the need to upskill the workforce in digital and AI skills (Cetindamar and Abedin, 2021; Cetindamar et. al 2022; Lokuge and Duan 2023). For example, in the last decade, the shift to cloud computing sparked demand for new professions such as Cloud Solution Architects and DevOps engineers, redefined the skills needed by others, for example Database Administrators and Networking Engineers,and reduced the need for professionals with on-premises skills like wiring and connecting hardware (Kotha et al. 2023). The rapid adoption of generative AI technologies has recently given rise to new roles such as prompt engineers and vector database developers.

Organisations wanting to maintain their strategic advantage must adapt to changing business conditions. This implies understanding future workforce needs and gaps in the current workforce supply to infer talent acquisition and learning and development (L&D) programs. Advanced machine learning approaches have been used to detect emerging labour market trends from the large quantity of real-time data posted online in job advertisements published on job portals and in professional resumes shared on professional social networks (Alabdulkareem et al. 2018; Althobaiti et al. 2022). However, there currently need to be more solutions that directly link strategic business initiatives - aimed at translating company goals and long-term vision into practice, such as adopting new technologies, expanding into new markets, or launching new products - to changes in labour demand.

This work addresses the abovementioned challenges by proposing a new ontology that explicitly links strategic initiatives to their role requirements. Several formal ontologies have been developed to link skills to occupations or to model the relationship between occupations and the different units within an organisation. However, to our knowledge, none has been developed to link roles to business initiatives that could span multiple organisational units and geographies. We call these business transformation initiatives workforce demand and supply drivers as they drive the demand for new roles and skills from employers and the supply of talent from job seekers. This study also outlines approaches and challenges in semi-automatically populating the ontology using a collection of real-time and historical data streams from online posting fora, such as Seek and Indeed.

We show that our approach to automatically link business transformation initiatives to job postings leveraging natural language processing is successful in returning either different specialist roles directly related to the adoption of a technology, e.g., cloud engineering roles of various description for cloud computing, or a broader range of occupations that could intuitively be linked to a business transformation initiative like strategy planners and marketing coordinators for entry in new markets. We demonstrate how our approach could be used in various scenarios to guide corporates and educational institutions on the workforce requirements needed when an organisation is transforming their business processes or adopting new technology.

The main contributions of this work are as follows:

- a novel ontology to link workforce demand and supply drivers to occupations and skills

- a novel discovery: job ads contain information about transformation initiatives that can be used to link different roles to the adoption of specific technologies or business initiatives

- an approach to leverage job ads as a data source for planning initiatives that relate to business transformation initiatives

# 2 Prerequisites and Related Work

## 2.1 Revisiting the Industry Standard

To maintain or gain competitive advantages, corporates must plan for the talent they need to execute their corporate strategy. This typically means hiring personnel or upskilling and reskilling human capital in areas where the business intends to expand and reducing resources in areas no longer considered business-critical.

Strategic workforce planning (SWP) is a proven approach to help organisations build their recruitment, learning and internal mobility plans (Gibson 2021). It has recently regained popularity since the fast pace of technological adoption has forced organisations to evolve and reshape their workforce composition continuously.

SWP has a longer time horizon than other workforce management approaches, such as tactical talent management, scheduling and rostering (Gibson 2021, Jaillet 2022). As such, it better accomodates transformative business initiatives spanning over several years and changes in the workforce composition.

SWP practices usually use qualitative and quantitative methods to estimate workforce requirements and turnover (Zhao et al. 2019; Safarishahrbijari 2018; Rowe and Wright 2001). However, forecasting at longer time horizons involves a higher degree of uncertainty as different factors might impact a company's staff requirements (Gibson 2021). For this reason, quantitative methods have been limited to a few specific applications.

There needs to be more alignment in using employee and labour market data to forecast workforce requirements. On the one hand, corporates use employee data to estimate turnover patterns, predict workforce supply needs and solve short-term scheduling problems. On the other hand, people analysts and education institutions use external labour market data sources to detect emerging trends and predict long-term workforce requirements. However, thus far, no attempt has been made to link specific strategic initiatives with the emergence of new jobs and skills.

We argue that text contained in job ads discloses information concerning the strategic initiative that drives the need for a role. Figure 1 exemplifies this hypothesis.

**Senior Software Engineer**

A global commercial software house is building a Blockchain product for a highly technical financial enterprise. This project has real world application impacting millions of real customers; Creating greater transparency, efficiency and value for all parties involved in financial trading.

**Head of Product - Retail**

This global retail analytics company is looking for a **Head of Product** who will play a leading role in building a brand new product, as well as strategically developing the entire function in a high growth organisation.

**Business Development Manager**

We have recently experienced significant growth in Australia and are about to launch internationally to provide flexible, affordable payment options to clients across multiple industries.

*Figure 1: Example of excerpts from job postings containing information on the strategic initiative driving the hiring (highlighted in pink boxes)*

The first ad mentions that the Senior Software Engineer position is available due to a strategic transformation initiative to develop a new product using Blockchain technologies.

In the second ad, the advertised role is related to the launch of a new product, whilst the third ad cites international expansion as the reason the job is available.

### 2.1.1 Taxonomies, Ontology and Databases

Although job postings and job seeker profiles provide significant sources of information, these documents do not share standardised formats, making it hard to analyse vast volumes of data. In addition, most of the information is in plain text and presents lexical ambiguities.

In the past few decades, Ontology-based Information Extraction (OBIE) methods have been used to improve the matching performance of Natural Language Processes (NLP) used to extract information from job ads (Sibarani at al. 2018; Terblanche et al. 2017).

This paper uses the following terminology to identify different technical solutions.

*Taxonomy* identifies a hierarchical and static representation of terms generally limited to one domain (Hedden 2010).

*Ontology* is a formalised knowledge representation generally of a specific domain. Concepts, properties, attributes, and their relationship are formally specified (Smith 2004). Compared to taxonomies, ontologies can extend to different domains, and the relationships between concepts can be dynamic, thus providing richer sources of information.

*Databases* are technology solutions designed to query and store data efficiently. They are typically designed to facilitate fast data processing, not to establish logical connections between concepts - although their schema's entities, attributes, relationships and constraints is defined. Databases are usually aimed at facilitating the fast processing of queries rather than at establishing logical connections between concepts, and typically embed less semantic meaning.

Leveraging ontologies allows for better capturing of the relevant terms of a knowledge domain and their relationships, thus helping with word disambiguation and analytical tasks. In addition, ontologies can be used to create a common language and facilitate data sharing. This is particularly useful in large corporate environments needing to consolidate data from multiple Human Resources Information Systems (HRIS).

## 2.2 Labour Market Ontologies

Several job-skill ontologies are applied to labour market data and formalise the relationship between occupations and their required skills (Sibarani at al. 2018; Terblanche et al. 2017; Djumalieva and Sleeman 2018). However, no existing ontology connects roles and skills to the business activities generating the demand for the roles needed within the organisation.

The digitisation of recruitment processes and the widespread adoption of online platforms and services created vast sources of real-time labour market data and information. These data sources have been used extensively to improve the match between employers and job candidates (Kircher 2020). Analyzing these novel data assets has allowed researchers to cast fresh light on labour market trends and dynamics (Dadzie et al. 2018). The continuous flow of job ads and candidate profiles posted online has also served as a leading indicator of hiring decisions and as a rich source of information on emerging skills in specific sectors (Galkin et al. 2018; De Mauro et al. 2018).

In recent years, these data sources have been combined with other datasets containing information on intellectual property patents, survey data on salary levels, occupation demographics and work typology, e.g., contract versus permanent work, remote work and flexible work arrangements,to analyse the highly dynamic nature of job markets. Examples include: estimating the impact of digital transformation and the automation of industrial processes on the labour market (Frey and Osborne 2017), job security, and minimum income levels (Faryna et al. 2022).

Job ads have also been used to study the polarisation between physical and cognitive jobs (Althobaiti et al. 2022; Alabdulkareem et al. 2018), to analyse urbanisation patterns (Ascheri et al. 2021) and the exposure of jobs to the Fourth Industrial Revolution (4IR), and to measure the fitness of higher education offerings to the requirements of the labour market (Ahadi et al. 2022) and recommending job transitions pathways (Dawson et al. 2021).

More recently, occupation titles were found to change less frequently than the skillset required to perform a job. A recent study (Althobaiti et al. 2022) found that only 1.67% of job titles disappear yearly. In other words, the job is labelled the same way, but the skills required to perform it can change significantly. Therefore, the focus of research activities has shifted toward detecting emerging skills and skill shortages (Dawson et al. 2021 ).

Most of the research efforts in this area leverage Natural Language Processes combined with Deep Neural Network approaches to extract information from job postings and forecast skill-set evolution in specific sectors. A few studies use OBIE methods to build clusters of skills and occupations (Sibarani and Scerri 2020) and quantify the demand for skills across industries or predict the evolution of skills based on word co-occurrence in job ads taken at a different point in time (Djumalieva et al. 2018; Sibarani and Scerri 2020). However, these analytical approaches rely on the frequency of specific skill terms within job descriptions. They need to consider the relationship with the business initiatives that might have determined the demand for skills. This has motivated our exploration of how this relationship can be leveraged in developing a new ontology.

## 3 Ontology Entities and Use Cases

This section presents a few use cases to demonstrate how our proposed ontology can be used to link occupations to business transformation initiatives and the entities and relationships represented.

### 3.1 Strategic Workforce Planning and Use Cases

In order to stay ahead in the market, companies must carefully evaluate and adapt their workforce and talent strategy. This includes planning for the necessary skills and expertise to effectively execute their business objectives.

In SWP, workforce requirements are defined in terms of capacity, the amount of workforce demand denoted in FTE, and capability, the skills, knowledge and abilities necessary to perform a job (Gibson 2021). Estimating workforce demand is the most challenging part of an SWP process, as several factors need to be considered. The proposed ontology would help analysts and planners identify the roles and skills required for corporate strategic initiatives.

Below, we present a few use cases to exemplify how companies could leverage an ontology of workforce demand drivers in the context of SWP.

*USE CASE 1 - New market entrance*

Company A plans to develop metaverse applications to improve its supply chain processes. They need to know the roles and skills required to start prototyping and build a Minimum Viable Product (MVP). They also need to plan for the roles and skills required to operationalise the solution and scale up the business segment once the commercial value of the solution has been proven.

*USE CASE 2 - New ways of working*

Company B is planning to adopt agile software development methodologies across their functions. They need to know the roles and skills required to initiate the change and the skills they need to develop across their workforce to champion adoption and ensure uptake.

*USE CASE 3 - Adoption of new technology to improve processes*

Company C is planning to digitise and automate their customer centre operations. They need to know what roles are required to kick-start the take up of the new technologies and initiate the transition. They must also know what skills to develop across their workforce to manage change and ensure technology adoption.

USE CASE 4 - *Maintain competitive advantage*

Company D has been experimenting with Quantum Computing for several years. They want to maintain their leadership position in the market and need to understand the emerging roles and skills in this field.

### 3.2 The Innovation-to-Occupations Ontology

We now define the entities within our proposed ontology and present their relationships. Figure 2 provides a graphical illustration of the entities listed in our ontology and the relationships between them.

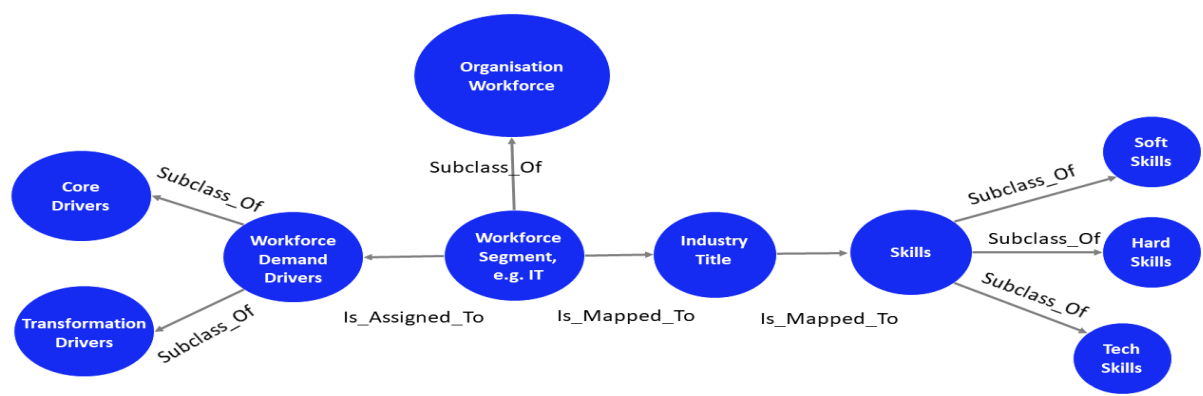

*Figure 2: An illustration of entities of the workforce ontology*

### 3.2.1 The Ontology Entities

Our ontology contains information about five core entity types.

1. *Organisation Workforce is* the entire workforce of an organisation.
2. *Workforce Segment* is a portion of the workforce of an organisation. In an SWP project, this might refer to the workforce of a specific geographic location, function, department, or any of their combinations.
3. *Industry Title* is the occupation title from a standard occupation classification mapped to the roles of an organisation. For instance, Junior Data Scientists, Senior Data Scientists, and Data Modellers could all be mapped to the Data Scientist Industry Title.
4. *Workforce Demand Drivers* are business activities, tasks or initiatives linked to one or more workforce segments.
5. *Skill*s are the set of knowledge, skills and abilities mapped to an Industry Title.

### 3.2.2 Connecting the Ontology Entities

Our ontology links Workforce Demand Drivers indirectly to Industry Titles via Workforce Segments. We make this design choice so that mapping roles within an organisation to standard occupation classifications is more straightforward. This would have two main advantages. Firstly, it would allow for re-using existing job-skill taxonomies where skills are linked to occupations within standard industry classifications. The link to skills would also allow to identify the competencies needed for specific business initiatives. Secondly, it could extend the ontology to traditional labour market concepts and data sources in future stages. The data points collected by government and statistical agencies are usually classified according to standard occupation taxonomies. Including these additional data sources would help companies infer their talent management and mobility plans. For instance, once a company has identified the roles and skills for adopting a specific technology, it might be interested in the average salary levels and employment prospects.

### 3.2.3 Core vs Transformation Demand Drivers

When estimating workforce demand, SWP considers activities that occur during an organisation's ordinary course of business, i.e., Business-As-Usual (BAU) activities, as well as strategic activities outside of BAU.

Our proposed ontology distinguishes workforce demand drivers into Core Business Drivers and Transformation Drivers.

*Core Business Drivers* are quantifiable business drivers of activities carried out by the organisation and its workforce as part of its regular course of business.

*Transformation Drivers* are strategic initiatives that occur outside of BAU.

Generating Core Business Drivers usually follows a bottom-up approach where key activities performed by specific occupations are analysed. Information about these activities may be contained in the list of job requirements of a job posting or could be available from job content databases such as O*NET. Some Core Business Drivers may be specific to certain occupations. For instance, the number of hospital admissions determines the number of nurses required in a hospital ward. Some others may be more generic and relate to multiple roles within an organisation, e.g., the volume of sales in a retail firm determines the number of staff required in business development and logistics departments.

Transformation Drivers more directly relate to the strategic initiatives of an organisation and the activities that occur outside of BAU. As previously shown, some examples of Transformation Drivers are the adoption of new technologies and new ways of working but also step change business initiatives such as the entry into a new market or the development of new products. Generating Transformation Drivers typically follows a top-down approach where the industry or cross-industry themes are analysed and linked to specific occupations. As we have seen, job postings detailing why a position is available can contain this information. Complementary qualitative research approaches such as nominal group technique, crowdsourcing and Delphi method could also be leveraged to identify specific emerging trends within or across industries (Gibson 2021).

## 4 Methodology

This section presents the approach followed to populate the ontology semi-automatically. Populating an ontology is often a resource-intensive process, and an approach must be found to extract the relevant information from vast job posting data.

### 4.1 Case study

For this case study, we focus on ten business transformation initiatives derived from industry reports on emerging technologies and society megatrends (Naughtin et al. 2022, World Economic Forum, and Frontiers Media S.A. 2023), five related to the adoption of new and emerging technologies – cloud computing, generative AI, quantum computing, blockchain and cybersecurity – and five related to broader business transformative processes – Agile ways of working, product customisation, move to renewable energy, entry in a new market and smart manufacturing. The justification for the number of transformation initiatives selected was to provide a number of examples that whilst not completely exhaustive could prove that the approach worked under different scenarios.

We aim to generate embeddings for the job descriptions published in job ads and the definitions of the transformation initiatives taken from Wikipedia articles. We leverage the text-embedding-ada-002 model to generate the job postings embeddings and use the same model to embed a search query using definitions taken from Wikipedia articles as input. We hypothesise that the words on Wikipedia pages could be found in job postings related to adopting a specific technology or a change in business processes or be semantically linked to the terms used in the ads to define the requirements of occupations linked to transformation initiatives. We then calculate the cosine similarity scores between the embeddings obtained from the input query and those obtained from the job postings to find the closest job titles matching a specific transformation initiative. Finally, we define a threshold to analyse the highest jobs, ranking by cosine similarity.

### 4.2 Data sources

The digitisation of recruitment processes has led to the generation of vast amounts of data published online. Compared to other more traditional data sources on occupations gathered through surveys, real-time data sources can provide a more accurate view of current trends in the job market. Despite providing current and granular information on job openings and skills, these data sources are highly heterogeneous. Documents posted online come, in fact, in a variety of digital formats: HTML, Word, PDF or Google Docs files, to name the most common. In addition, the information in these documents does not follow standardised templates, is inconsistent and has varying levels of detail. To be analysed, it needs to be extracted, standardised and consolidated.

The primary data sources for real-time labour market information are:

- Job advertisements (Dawson et al. 2021; Ahadi et al. 2022) – also known as job ads or job postings - are online announcements to advertise an opening for newly created roles or fill vacant positions. In some cases, job ads are used to attract expressions of interest for roles within a company that could be opening in the future. The initial posting may contain text detailing why a job is available, which in turn might contain information on strategic initiatives and mention specific projects or technologies.
- Resumes (Frank et al. 2019; Djumalieva et al. 2018) – also known as curriculum vitae (CVs) - are formalised documents generally used to apply for vacancies. They often provide details of previous working experiences and achievements, including participation in strategic initiatives and transformative processes. These documents also often contain granular lists of job seekers' knowledge, skills, and abilities.
- Candidate profiles (Rajkumar et al. 2022) – are more succinct summaries of workers' abilities and past working experiences. They are usually posted publicly on professional platforms and are less formalised than resumes. A candidate profile's working experiences and achievements sections sometimes mention participation in specific projects and strategic initiatives.
- Ontologies (Boselli et al. 2018; Giabelli et al. 2020; Sibarani et al. 2018; Terblanche et al. 2017) – Real-time labour market data comes in various formats that must be pre-processed before being analysed. In recent research studies, standard occupational classifications developed by government agencies or third-party vendor occupations and skill taxonomies have been used to assist with standardisation and consolidation tasks. Research studies have also leveraged sector-specific knowledge bases to build and enrich occupation and skill ontologies.

For our case study, we use a dataset of 1,500,000 job postings scraped from various job portals (Indeed, LinkedIn, Seek) and published in Australia, the US, and the UK from March to June 2023.

## 5  Experiment and Results

We start by converting the original LD-JSON files into CSV format. We then perform initial cleansing to filter out duplicates and remove records incorrectly processed when parsing the LD-JSON files. Removing duplicates leads to 1,091,642 records. After removing records incorrectly parsed from LD-JSON to CSV format, we are left with 1,077,034 job descriptions. Before processing the records, we perform additional cleansing to the text of the job descriptions to remove punctuation, incorrect characters and multiple spaces to generate good-quality embeddings.

### 5.1  Implementation choices

We leverage the OpenAI services in Microsoft Azure to analyse the data and generate the embeddings using the text-embedding-ada-002 (Version 2) model. The model provides numerical representations of concepts by converting text into numerical sequences that can then be used to calculate the similarity between concepts contained in text. We chose this model as it has been shown to outperform all other OpenAI models in terms of text search and sentence similarity (https://openai.com/blog/new-and-improved-embedding-model). In addition, it is a more suitable model to work with longer portions of text. OpenAI documentation reports an increased context window from 2,084 input tokens to 8,192 for this version of the model, making it a better choice for longer documents. Accordingly, in our experiment, we have limited the input tokens to 8,192 to ensure that none of the data we pass to the model for tokenisation and embedding exceeds the input token limit for GPT-4 input files.

After processing the job posting records and Wikipedia definitions to obtain the embeddings, we apply a cosine similarity function to retrieve the job titles that more closely match the selected business transformation initiatives. A cursory examination of the results thus obtained suggests that a similarity score threshold of 0.70 should be adopted to limit the analysis to job titles that more closely match a business transformation initiative.

### 5.2  Experiment Results

Figure 3 shows the distribution of the job titles matching each of the selected business transformation initiatives.

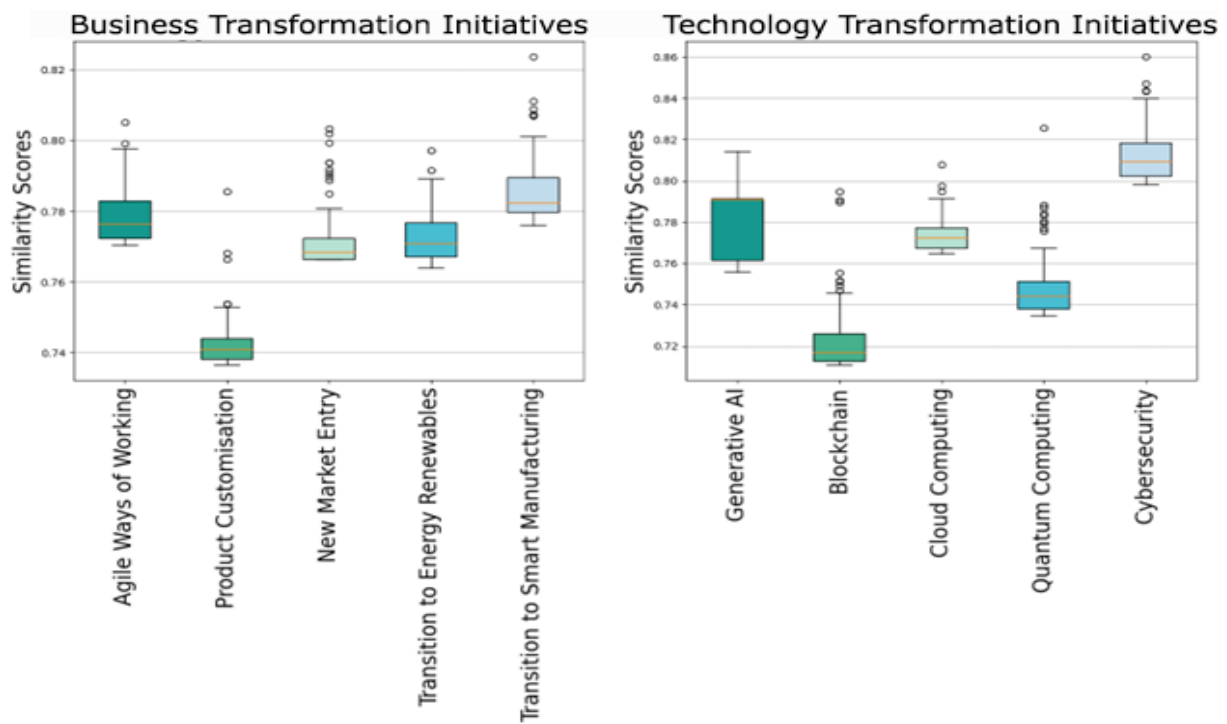

*Figure 3: Frequency distribution of job titles more closely matching business transformation initiatives by similarity score*

The results show that our approach returns either specialist roles linked to the roll-out of a specific technology, such as cloud engineering roles of various description for Cloud Computing or a broad range of occupations intuitively related to a business transformation initiative such as strategy planners and marketing coordinators for Market Entry. Figure 3 shows high similarity scores for all business transformation initiatives but somewhat lower similarity scores for Product Customisation and the adoption of Blockchain and Quantum Computing. For Product Customisation, we note that the words used in the job description are less semantically related to the term used in the Wikipedia article as input for the query. However, the routine still performs well, retrieving marketing or product development roles. For blockchain adoption, the lower similarity scores could be justified because the word "blockchain" or other words usually used to describe the technology less often appear in job titles and job descriptions. However, looking closely at the top-ranking occupations for blockchain adoption, we note that most occupations retrieved are openings within firms whose core business is peer-to-peer payment systems, cryptocurrency products or other blockchain applications. Hence, the results are still valid. Looking closely at the text of the job descriptions retrieved for Quantum Computing, we note that the word quantum is sometimes used to convey an idea of sophistication and is not related to the adoption of quantum technologies. This distorts the results, returning high similarity scores even when the occupations are not directly linked to quantum computing. However, this issue appears to be limited to a few cases.

We then look more closely at the list of top 10 roles ranking by similarity score. The results for business transformation initiatives are reported in Appendix 1, while those for transformation initiatives linked to technological adoption are reported in Appendix 2. The lists have been constructed by consolidating job titles identifying the same occupations. For instance, Senior Cloud Engineer, Lead Cloud Engineer and Cloud Engineer roles would be assimilated into the Cloud Engineers category. Job postings advertising the same occupation in different locations have been counted once to avoid double-counting. The list shows that our approach performs well, returning a variety of roles that are either explicitly connected with a specific technology or a broad range of roles that could intuitively relate to the business transformation initiative under consideration. The results also show how our approach is successful in linking a variety of roles through semantic similarity. For instance, for Generative AI, our approach retrieves specialist roles such as Post-Doctoral Researchers and Deep Learning Engineers but also more generalist roles such as Software Developers, Architects and Sales Executives involved in the operationalisation and commercialisation of Generative AI solutions.

## 6   Conclusions

Our research demonstrates that our approach to mapping business transformation initiatives to job titles using natural language data and OpenAI technologies produces intuitive results for most transformation initiatives under consideration. We show that our methodology allows us to retrieve either specialist occupations directly linked to the adoption of specific technologies, such as cloud engineering roles of various description for Cloud Computing or a broader set of occupations that can intuitively be related to business transformation initiatives, such as strategy planners and marketing coordinators for entry in new markets. The current approach however presents some limitations. Firstly, it relies on Wikipedia articles covering business transformation initiatives being published. Whilst we could find a good selection of articles covering different business transformation initiatives, articles for emerging transformative processes might not have been published yet. Furthermore, to our knowledge, there isn't a fully exhaustive term list or taxonomy of business transformation initiatives that could be used as a benchmark to validate our approach against a comprehensive set of scenarios. Finally, although we have shown that the information on business transformation initiatives can be found in job postings, not all job ads would contain it. Short-term and low-skilled vacancies, for instance, are less likely to contain this information. The results presented in this paper are ongoing, and as such further improvements could be achieved by leveraging a larger sample of job ads, using larger corpora as user input in the search query or obtaining better definitions for business transformation initiatives from other sources, e.g., industry white papers, research articles, practitioner blog articles or focus group of subject matter experts. As the focus of talent management practices has recently been around the skills within specific occupations or occupation categories, the approach proposed in this paper could also be used to identify core skills within specific transformation initiatives, linking the information on skills contained in job postings to Wikipedia articles describing business transformation initiatives.

## Appendix 1

| Cybersecurity | Quantum Computing | Cloud Computing | Blockchain | Generative AI |
| --- | --- | --- | --- | --- |
| Cybersecurity Specialists | Staff Engineer Opto-Mechanicals | Cloud Service Admins | Blockchain Technical Analysts | Software Architect - AI/ML/Clouds |
| IT Security Analysts | Plasma Etch R&D Engineers | Cloud Service Engineers | Content Marketing Managers | Quality System Directors |
| Information Technology Security Consultants | Cyber Engineers | Cloud Security Engineers | Blockchain Engineering Managers | Artificial Intelligence Consultants |
| Cybersecurity Managers | Cybersecurity Operation Compliance Leads | Cloud Operations Engineers | Social Value and Sustainability Leads | OpenAI/GPT3 Experts / AI/ML Developers |
| Security Administrators | Enterprise Account Executives | Product Managers - Private Cloud | Head of Engineering | Post-Doctoral Scientists - AI & Machine Learning |
| Information Security Admins | Power Systems Advisors | Principal Engineers | Data Scientists | Software Architect (AI/ML) |
| Security Analysts | Cybersecurity Operations Analysts | Cloud System Engineers | Blockchain Engineers | Enterprise Sales Executives |
| Cybersecurity Analysts | Computer Scientists | Cloud Solution Architects | Managers - Engineering | Deep Learning Engineers |
| Cybersecurity System Engineers | Portfolio Managers | Cloud Engineers | Software Engineers | Data Collection |
| Cybersecurity Architects | Research Associates | Cloud Activation Managers | User Researchers | Director, Drug Product & Device Developments |

## Appendix 2

| Agile Ways of Working | Customisation | Market Entry | Smart Manufacturing | Transition to Energy Renewables |
|---|---|---|---|---|
| Business Analysts | Director Custom Research | Marketplace Specialists | Manufacturing Engineers | Project Coordinator - Renewable Systems |
| Enterprise Business Analysts | Director Product Management | Strategy Planners | Manufacturing Planners | Estimators – Sustainable Energy |
| SQL Database Administrators | Product Engineering Specialists | Marketing Coordinators | Industrial Engineers | Sustainability and Commercial Development Specialists |
| Agile Business Analysts | Team Leads | Marketing Managers | Maintenance Specialists | Protection Engineers - GHD |
|  | Tailors/Seamstress | Import Agents | System Engineers | Energy Innovation Analysts |
| Business System Analysts | Lead Functional Consultants | Quoting Specialists | Manufacturing System Analysts | Salespersons |
| IT Business Analysts | Product Managers | Management Consultants | Statistical Process Control Managers | Procurement Directors, Renewable Energy |
| Business Managers | User Researchers | Marketing Assistants | Project Managers | Energy and Sustainability Project Managers |
| Strategic Capability Partners | Manufacturing Execution Systems | Area Sales Managers | Manufacturing Reliability Managers | Network Modelling and Renewable Energy Engineers |
| Managers | Account Executives | Export Operations Specialist | Site Managers | Safety, Health and Environmental Apprentices |